\documentclass[letterpaper, 10 pt, conference]{ieeeconf}  

\IEEEoverridecommandlockouts                              

\usepackage{amsmath, amssymb , graphicx, subcaption}
\usepackage{multirow}
\usepackage{hyperref}
\usepackage{dblfloatfix} 
\usepackage{balance}
\usepackage[symbol]{footmisc}

\usepackage{xcolor}

\title{\LARGE \bf 
ARCSnake: An Archimedes' Screw-Propelled, Reconfigurable Serpentine Robot for Complex Environments}

\author{Dimitriw A. Schreiber$^{\dag, 1}$ \IEEEmembership{Student Member, IEEE}, Florian Richter$^{\dag,1}$ \IEEEmembership{Student Member, IEEE},\\ Andrew Bilan$^{\ddag,2}$, Peter V. Gavrilov$^{\ddag,2}$, Hoi Man Lam$^{\ddag,2}$, Casey H. Price$^{\ddag,3}$, \\Kalind C. Carpenter$^4$, and Michael C. Yip$^1$ \IEEEmembership{Member, IEEE}
\thanks{$^\dag, ^\ddag$Equal contribution}
\thanks{$^1$Department of Electrical and Computer Engineering, University of California San Diego, La Jolla, CA 92093 USA. {\tt\small \{dschreib, frichter, m1yip\}@ucsd.edu}}
\thanks{$^2$Department of Mechanical and Aerospace Engineering, University of California San Diego, La Jolla, CA 92093 USA. {\tt\small \{abilan, pgavrilo, hml024\}@ucsd.edu}}
\thanks{$^3$Department of Computer Science and Engineering, University of California San Diego, La Jolla, CA 92093 USA. {\tt\small c2price@ucsd.edu}}
\thanks{$^4$NASA Jet Propulsion Laboratory, Pasadena, CA 91109 USA {\tt\small kalind.c.carpenter@jpl.nasa.gov}}
}
\begin{document}

\maketitle
\thispagestyle{empty}
\pagestyle{empty}
\begin{abstract}
This paper presents the design and performance of a new locomotion strategy for serpentine robots using screw propulsion. The ARCSnake robot comprises serially linked, identical modules, each incorporating an Archimedes' screw for propulsion and a universal joint (U-Joint) for orientation control. When serially chained, these modules form a versatile serpentine robot platform which enables the robot to reshape its body configuration for varying environments, typical of a snake. Furthermore, the Archimedes' screws allow for novel omni-wheel drive-like motions by speed controlling their screw threads.
This paper considers the mechanical and electrical design, as well as the software architecture for realizing a fully integrated system.
The system includes 3$N$ actuators for $N$ segments, each controlled using a BeagleBone Black with a customized power-electronics cape, a 9 Degrees of Freedom (DoF) Inertial Measurement Unit (IMU), and a scalable communication channel over ROS. 
This robot serves as the first proof-of-concept demonstration of the NASA-JPL Exobiology Extant Life Surveyor (EELS) program that aims to deliver scientific instrumentation deep within the plume vents, caves, and ice sheets of Enceladus and Europa in search for extant lifeforms\footnote[1]{Website: \fontsize{7}{8}{\url{https://www.sites.google.com/ucsd.edu/arcsnake}}}.
\end{abstract}

\section{Introduction}


\begin{figure}[t!]
    \vspace{2mm}
    \centering
    \includegraphics[width=\linewidth]{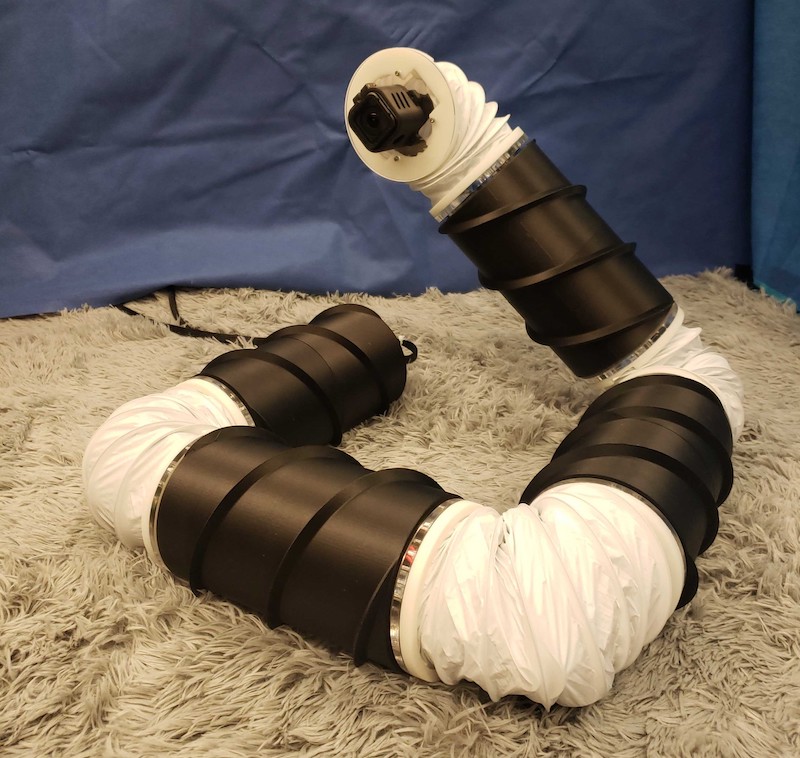}
    \caption{ARCSnake is a screw-propelled, serpentine platform for reconfigurable locomotion in complex environments.}
    \label{fig:u_joint}
    \vspace{-5mm}
\end{figure}

Snake-inspired robots are multi-segmented hyper-redundant robots comprised of repeated joint-segment units. 
They are kinematically flexible and low profile, which allows them to conform to their environment and enter confined terrain.
Therefore, they can explore environments which preclude traditional robotic systems.
 Snake-inspired robots can be classified into two main categories \cite{granosik2005integrated}: \textit{serpentine} and \textit{snakelike} robots.
Serpentine robots propel themselves using active skins, and their joints are either passive or actively controlled to better position the links for more optimal grip.
In contrast, snakelike robots have passive skins and active joints to propel themselves.

{
\renewcommand{\arraystretch}{1.0}
\begin{table}[t]
    \vspace{2mm}
    \label{tab:specs}
    \centering
    \begin{tabular}{c|l|l}
    \textbf{Specifications} & \textbf{Within} & \textbf{Values} \\ \hline
    \multirow{2}{*}{Power} & Module & 12-60V, 310W (max)   \\ 
     &System & 12-60V, 1240W (max)\\ \hline
    \multirow{2}{*}{Communication} & Module & I2C \\
    &System & TCP/IP via ROS \\ \hline
    \multirow{6}{*}{Sensors} &  \multirow{4}{*}{Body} & Optical Encoder\\
                          && Motor Current\\
                          && 9 DoF IMU\\
                          && Temperature Sensor\\
     & \multirow{2}{*}{U-Joint} & Magnetic and Optical Encoders\\
     && Motor Current\\ \hline
     \multirow{2}{*}{Actuators} & Screw & Torque: 1.6Nm continuous, 2.0Nm peak\\
     & U-Joint & Torque: 2.1Nm continuous, 2.7Nm peak\\ \hline
      \multirow{3}{*}{Dimensions} & Body & Max Len: 19.6cm \  Max Dia: 12.5cm\\
      & U-Joint & Max Len: 16.8cm \  Max Dia: 11.0cm\\
     & System & Max Len: 128.7cm Max Dia: 12.5cm\\   \hline
     \multirow{3}{*}{Mass} & Module & Body: 1.0kg, U-Joint: 0.88kg\\ 
     & Head & 0.68kg \\
     & System & 6.1kg \\
    \end{tabular}
    \caption{ARCSnake design specifications highlighting the redundant sensing and kinematic chain. A Body contains an inner shell and an Archimedes' screw.}
\end{table}
}

In this paper, we examine the design of a novel serpentine robot which we call ARCSnake, focusing on our application of the Archimedes' screw as an active skin for terrestrial propulsion. Mechanically, each segment is composed of a powered 2 Degrees of Freedom (DoF) universal joint (U-Joint) and an Archimedes' screw. 
This spherical joint rotation with high torque output enables flexible placement of the Archimedes' screw. Therefore, it allows reconfigurable drive patterns that transition from forward tunneling to various omni-wheel drive configurations.
Furthermore, careful implementation of robust software and electrical architectures reinforce the robot's exploration capabilities in unknown environments and will enable future tether-free operation. 

ARCSnake is the first proof-of-concept robot platform for the NASA-JPL Exobiology Extant Life Surveyor (EELS) program whose mission objective is to dive into kilometer-long plume vents, crevices, and ice caves on Enceladus and Europa, moons of Saturn and Jupiter respectively, to access subterranean oceans in search for extant life forms. The capability to efficiently navigate and perch in tortuous terrain in tight vertical passages such as vents and crevaces, while also effectively navigating open spaces such as ice caverns, motivates the screw-based design and its many locomotion capabilities. 
\section{Related Work}

\begin{figure*}[b]
    \centering
    \includegraphics[width=1\linewidth]{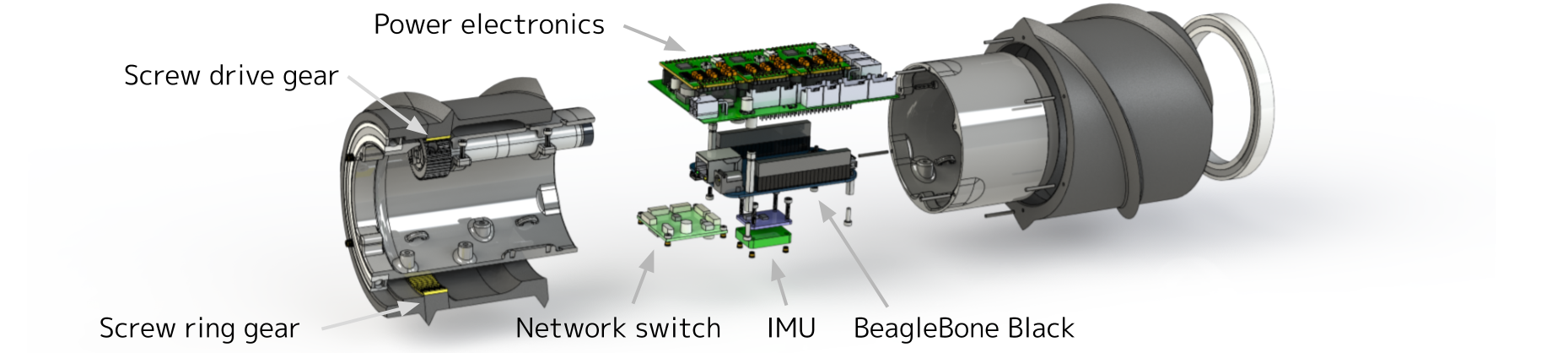}
    \caption{An exploded view of a body from the proposed robotic serpentine design. The body uses an Archimedes' screw as an active skin for unique locomotion capabilities while remaining mechanically simple. Inside the body is an embedded system that runs actuator controllers locally for easy scalablility.}
    \label{fig:exploded_view}
\end{figure*}

Designs for snake robots have been considered for decades \cite{hirosetutorial}, and have followed trends in the development of motor technology for shaping its body. We discuss recent examples in regards to their gaits with respect to serpentine or snake-like behaviors, and considerations of their mechanical and software design for adaptability to complex and varied environments. 

In general, using body contortion for locomotion is less efficient for traveling, but has inherent advantages in controlling body configuration with respect to rough or climbing terrain. The Unified Snake \cite{wright2012design} and its successor the SEA Snake \cite{rollinson2014design} comprise repeated perpendicularly offset joints, and each link involves a custom series elastic actuator, which incorporates both onboard control and scalable interfacing. The series elasticity allows safer conformation to its environment and force control capabilities as required for perching and climbing. However, offset joints make control a challenge, as maintaining a continuous line of contact with the ground or climbing surface is mechanically and kinematically non-trivial. 


The Amphibot \cite{crespi2005amphibot} \cite{crespi2006amphibot} undulates on the water's surface by yawing its joints to form a sinusoidal wave pattern.
Its passive wheels allow it to replicate the swimming motion on land, but only on planar surfaces, limiting its locomotive capabilities to only structured and flat environments. The ACM-R5 \cite{chigisaki2005design} amphibious snakelike robot utilizes 2-axis geared U-Joints protected by bellows. These joints have a $90^\circ$ range of motion on both axes. 
ACM-R5's segments are lined with passive wheels and fins, which provide anisotropic resistance on both land and water for propulsion. 
Perambulator \cite{5553836} branches off from the ACM-R5's design, implementing a 3-axis geared U-Joint housed within the length of the segment, while also using the same wheel-fin design as the ACM-R5. The design uses a differential drive mechanism, resulting in a roll about the central axis of the snake backbone; this leaves the robot in a singular condition when the robot is straight, which is undesirable in many situations where the robot may be required to snake its way into a tight space. 


Alternatively, serpentine robots that leverage an active skin mechanic for travel can realize efficient locomotion. Examples include the OmniTread \cite{borenstein2007omnitread}, which utilizes moving tracks along the length of each segment to drive along terrain. OmniTread's active skin provides robust terrestrial movement thanks to its high profile grip. Similarly, a toroidal skin drive was proposed which uses an active skin across the entire robot \cite{mckenna2008toroidal}. This active skin enables the robot to traverse complex spaces given its soft film-like exterior. However, the design is complex and therefore challenging to seal from particulates and fluids. The toroidal skin also provides forces tangential to the body configuration, which limits the locomotion capability of the skin to only the curve of the snake's body. At a larger scale, ACM-R8 \cite{komura2015development} utilizes large wheels with swing-grouser grips, flexing its body using hydraulics to maintain optimal grip while climbing obstacles.
All of these active skins, however, have complex mechanical designs and present complex control challenges.

The Archimedes' screw presented in this paper aims to maneuver the robot through complex terrestrial environments through a mixture of serpentine motions and snake-like configurations. Screw-type active skins are a new control paradigm for serpentine robots. Prior work by  \cite{fukushima2012modeling} shows the concept in 2D, where the robot uses passive wheels organized along a screw to reduce friction and developed a front-unit-following control method. The following sections describe a complete mechanical and software architecture realizing a screw-propelled, spherically jointed, serpentine robot. 

\section{Mechanical Design} 

Each module is composed of a self-contained 2-DoF U-Joint for positioning and an Archimedes' screw for propulsion. The U-Joint allows for $180^{\circ}$ rotation about each principal axis. Our system is composed of four such modules with a High Definition wireless camera attached to the head. The kinematic model of a single module is shown in Table \ref{tab:dh_params} using modified Denavit-Hartenberg parameters \cite{corke2007simple}.

{
\renewcommand{\arraystretch}{1.3}
\begin{table}[t!]
    \vspace{2mm}
    \centering
    \setlength\tabcolsep{1.1em}    
    \begin{tabular}{c|c|c|c|c}
    \textbf{Coordinate Frame} & a & $\alpha$ & D & $\theta$\\ \hline
    U-Joint Pitch & 28.0 & $-\frac{\pi}{2}$ & 0 & $q_p$ \\
    U-Joint Yaw & 0 & $\frac{\pi}{2}$ & 0 & $q_y$\\
    Next Module & 8.4 & 0 & 0 & 0\\
    \end{tabular}
    \caption{Kinematic model of a single module using the modified Denavit-Hartenberg parameters, where the joint angle positions are $q_p$ and $q_y$ for pitch and yaw respectively and the units are in centimeters and radians for distances and angles, respectively.}
    \label{tab:dh_params}
\end{table}
}


\subsection{Archimedes' Screw Design}
The Archimedes' screw, shown in Fig. \ref{fig:exploded_view} provides propulsion for the robot and allows for novel snake-like locomotion. A helix angle of $22^\circ$ was selected as it provided the greatest drawbar pull with the least amount of slippage \cite{archimedes_ref}. The optimal helix angle is affected both by the diameter of the screw and the pitch. The screw was designed to provide high maximum forces.
The two-start screw design was selected for increased soil contact. The screw has a root diameter of 112.5mm and an outer diameter of 128mm. This results in a 137mm helical pitch. 

A ECXSP16L motor with GPX16HP 35:1 gearhead drives the Archimedes' screw. This motor is capable of extremely high speeds that far exceed the gearbox's rated input speed. 
Therefore, the motor's speed is electronically limited to 12,000RPM. 
With the integral pinion-ring gear (3.4:1 ratio) in the screw and the transmission inefficiencies the screw is capable of 100 RPM, 1.6Nm continuous, and 2.0Nm peak torque. This results in a 0.23m/s screw lead speed and is an upper bound on the robot's maximum translational speed.

\subsection{Inner Shell}

The inner shell links to the Archimedes' screw and houses the electronics and propulsion motor. The screw and its corresponding inner shell are coupled through sealed thin profile bearings (VXB Part Num. 61816-2RS1) and are designed to house a 55Wh 5-cell Lithium-ion battery pack and support hermetic sealing. These extensions will be explored in future work.

\subsection{U-Joint Design}
The U-Joint, shown in Fig. \ref{fig:u_joint}, enables high dexterity with a large working range and minimal complexity. 
The joint is symmetric and composed of two identical halves, which are each driven by a Maxon ECXSP22M brushless motor with Maxon GPX22 44:1 low-backlash gearheads. Similar to the Archimedes' screw motor, this motor is capable of extremely high speeds that far exceed the gearbox's rated input speed. Therefore, it is also electronically limited to 12,000RPM. 

The motor is coupled to the U-joint through a 6mm width GT2 timing belt reduction (ratio 3.125:1). This timing belt offers very low backlash, high load rating, high stiffness, and fair transparency. A cable drive was considered as an alternative. However, it was forgone due to the simpler design and assembly of belt drives. After gearing, the joint is capable of moving at 87RPM. The joint's continuous output torque is 2.1Nm and peak torque is 2.7Nm after accounting for all transmission losses.  The 18N preload on the timing belt provides a 1.0Nm input pulley ratcheting torque, which is higher than our transmission observes during normal operation. The joint's torque is gearbox limited and could be increased, if necessary, with an alternate gearbox.

\begin{figure}[t!]
    \vspace{2mm}
    \centering
    \includegraphics[width=0.98\columnwidth, trim={0 0 0 10mm}, clip=true]{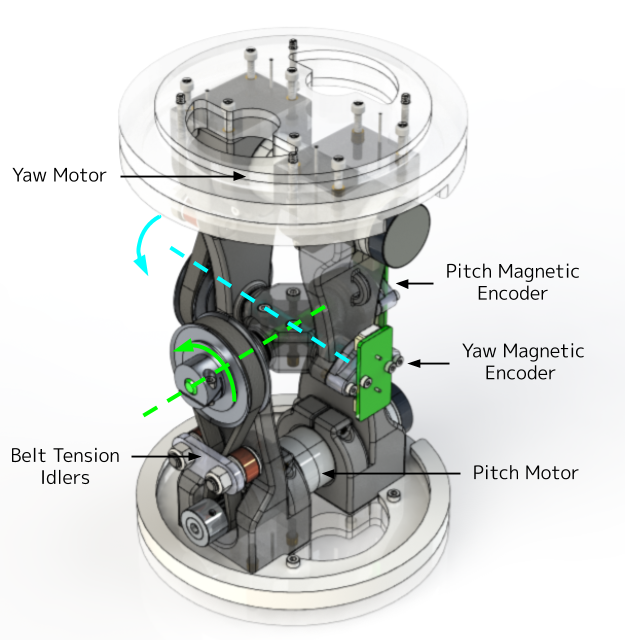}
    \caption{A render of the U-Joint connecting bodies of the proposed serpentine robot. Timing belts with tension idlers are used for transmission because of the simple mechanical design, high load rating, and low backlash.
    Magnetic encoders are placed on the axis of rotation for absolute position feedback and redundant sensing.}
    \label{fig:u_joint}
\end{figure}

\begin{figure*}[t!]
    \vspace{2mm}
    \center
    \includegraphics[width=1\textwidth]{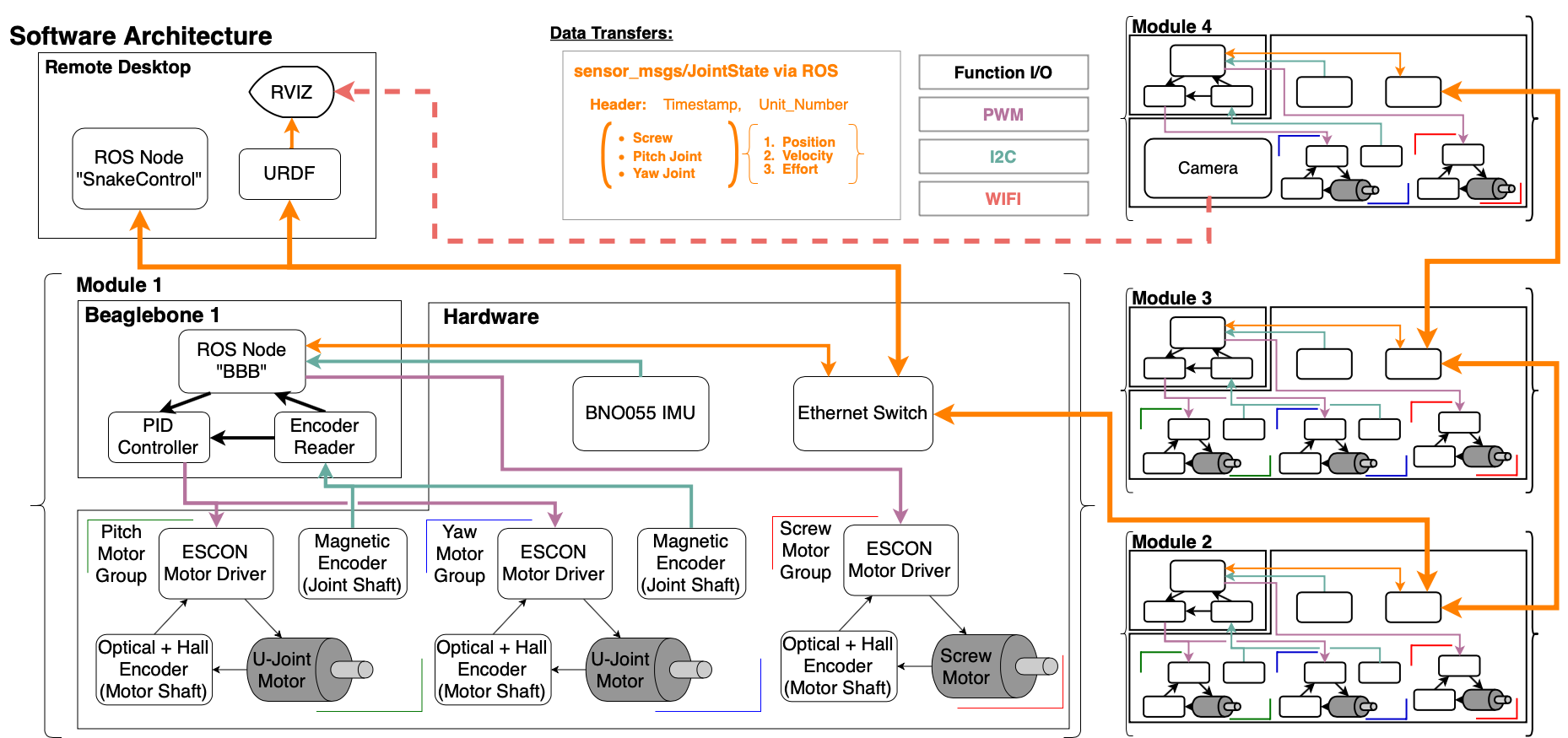}
    \caption{Software architecture for the modular serpentine robot. The embedded systems and a remote desktop communicate actuator state information with one another over TCP/IP via ROS. With this design, additional modules can be easily added.}
    \vspace{-3mm}
    \label{fig:flowchart}
\end{figure*}

\subsection{Manufacturing}
The robot's hardware was manufactured using a combination of 3D printing, laser cutting, and machining. 3D printing was used for the majority of parts to allow easier prototyping. Machined parts were only used in areas with very high stresses. Markforged FDM printers were used for structural components due to their ability to print parts reinforced with continuous fiberglass. Formlabs Form2 SLA printers were used for complex non-structural components. Simple flat structural components, such as the Archimedes' screw to U-Joint coupling plate, were laser cut from acrylic. The U-Joint crossbars were machined and brazed from 17-4 PH Stainless Steel rods.

\subsection{Sensors}
Two AS5048B I2C 14-bit magnetic encoders are placed on the U-Joint, as seen in Fig. \ref{fig:u_joint}. These provide absolute sensing of the rotary axis with $0.022^\circ$ resolution. The optical encoders on the motors provide redundant sensing to detect slipping and sensor failure.

A BNO055 Inertial Measurement Unit (IMU) is placed in each screw link to provide a world reference orientation. This 9 DoF System-in-Package IMU includes a 32-bit microcontroller which performs sensor fusion onboard and provides the filtered output. These are placed on separate breakouts, which are mounted distantly from motors with vibration dampening rubber.

\section{Electrical and Software Design}

Each body has its own embedded system for scalability. Minimal effort is required to attach additional units due to the design's communication architecture. The system is robust against voltage sags and signal integrity issues from long cable runs. 
A flowchart of the full electrical and software system is shown in Fig. \ref{fig:flowchart}.

\subsection{Electrical Design}

\begin{figure*}[t]
\centering
\vspace{2mm}
\includegraphics[clip=true,trim={5cm 0 5cm 0},height=1.33in]{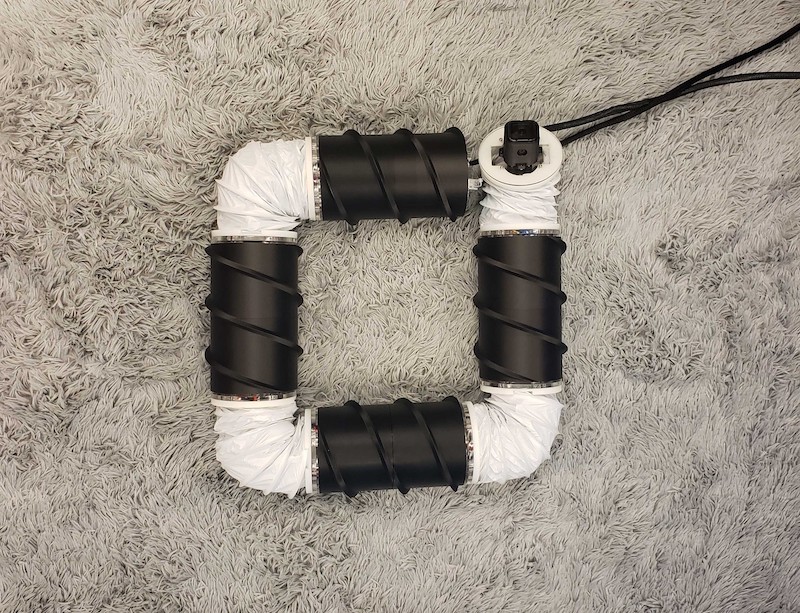}
\includegraphics[clip=true,trim={0 0 0 0},height=1.33in]{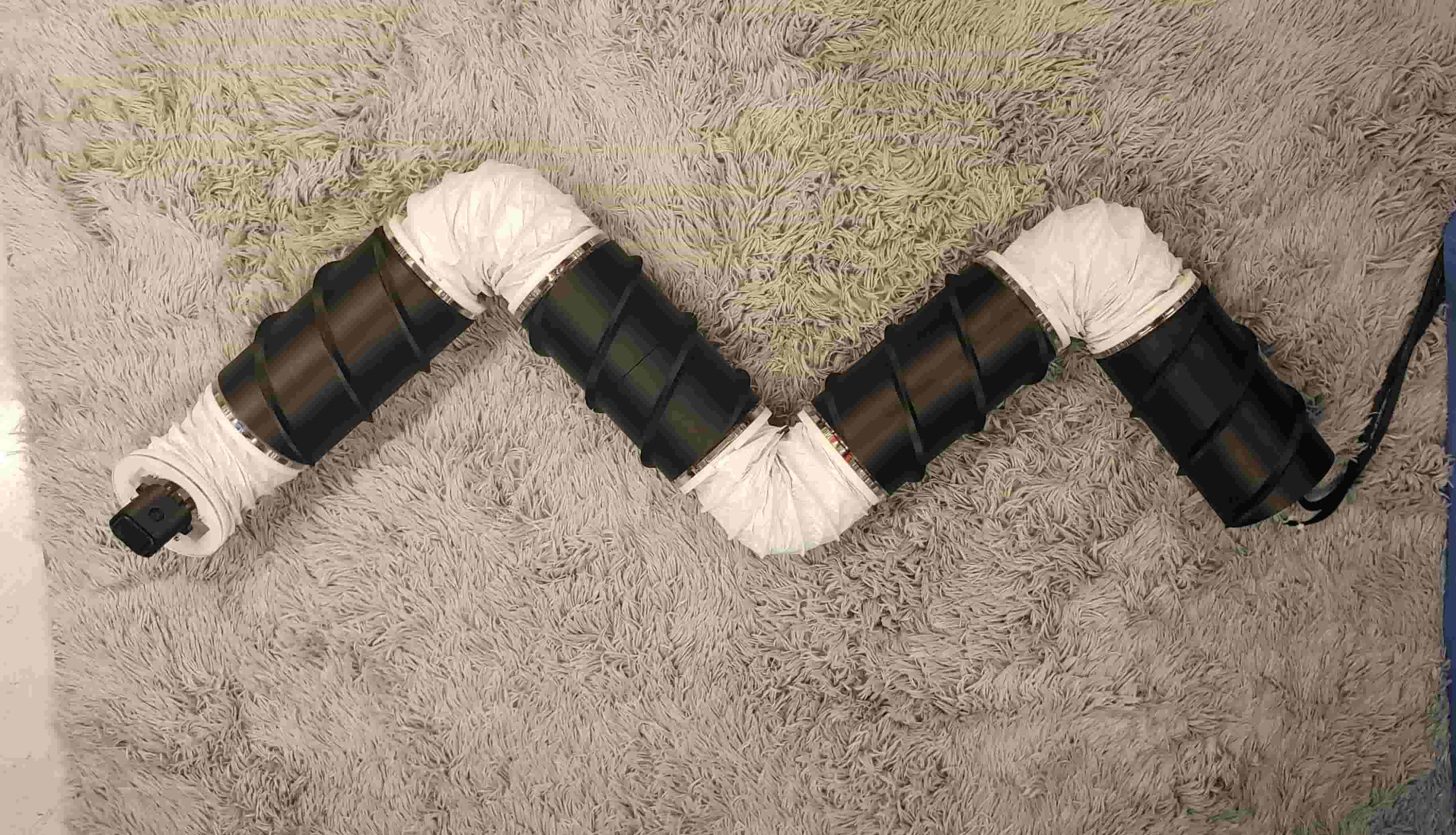}
\includegraphics[clip=true,trim={0 0 0 0},height=1.33in]{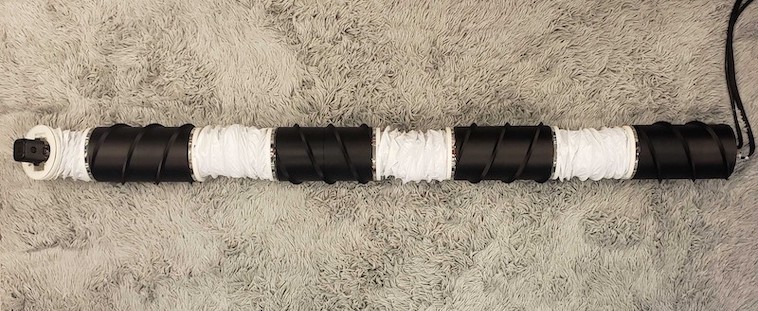}
\\
\includegraphics[clip=true,trim={5cm 0 5cm 0},height=1.33in]{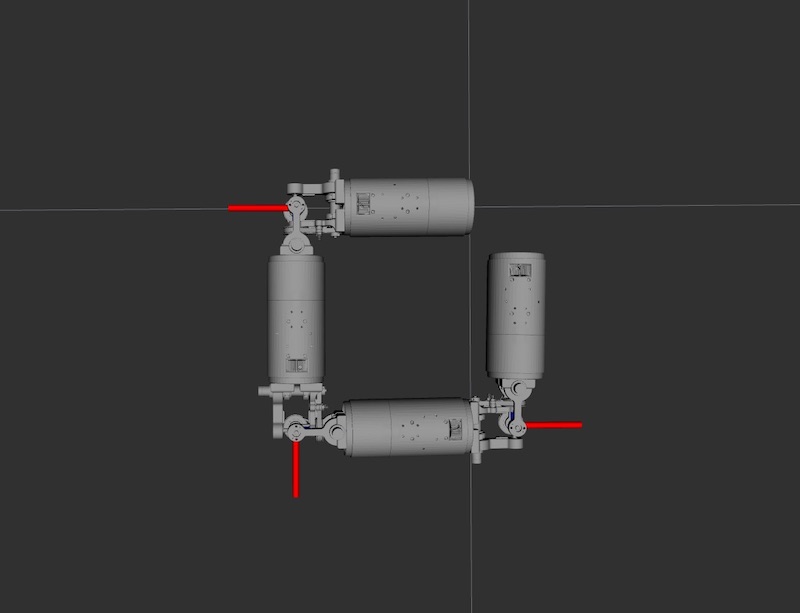}
\includegraphics[clip=true,trim={0 0 0 0},height=1.33in]{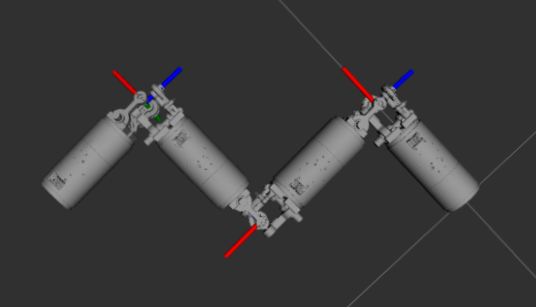}
\includegraphics[clip=true,trim={0 0 0 0},height=1.33in]{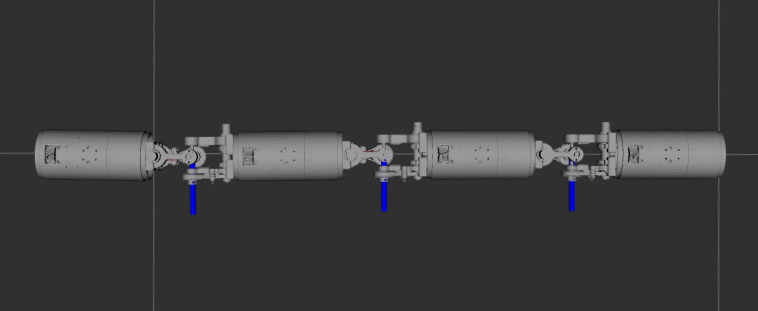}
\caption{Control and visualization in Rviz of various configurations on ARCSnake that shows its capability to conform its body to support different modes of locomotion.}
\label{fig:gaits}
\vspace{-3mm}
\end{figure*}

A BeagleBone Black \cite{beagle_bone} and custom cape is in every body. 
Three high efficiency buck-boost regulators, VICOR PI3740-00, power each cape's 24V rail and accept input voltages from 12V to 6V. 
The buck-boost regulators are connected in parallel and have a maximum output current of 10.2A to 12.7A, depending on the input voltage. 
The 24V power rail supplies the motor drivers. A separate step-down converter, Texas Instrument's TPS54561, provides 5V power to the BeagleBone Black and a low dropout 3.3V regulator, Texas Instruments TL5209, provides power to the ancillary components (magnetic encoder, IMU, and network switch). 
These regulators make the system robust against voltage sags on the power input and compatible with batteries in future iterations. 
Another benefit is that their many voltage, current, and temperature protections reduce potential failures such as back electromotive force from the motors.

Three Maxon Motor ESCON 50/5 brushless motor drivers are attached to the cape.
The motors have Hall effect sensors and optical encoders, which are directly connected to their respective motor drivers. Therefore, the motor drivers have the option of closed loop torque and velocity control.
Currently, all motors are set in velocity control mode and the current readings are sent via an analog signal to the BeagleBone Black.
The motor drivers have multiple built-in safeties, including temperature, voltage, and current protections, to ensure safe operation of the motors.

The ancillary components are located on separate breakouts.
The magnetic encoders (AS5048B) and IMU with temperature sensor (BNO055) both communicate with the BeagleBone Black via I2C.
The network switch (IP175G) connects the incoming and outgoing Ethernet to the segment and the BeagleBone Black. 
To improve signal integrity and for galvanic isolation when operating on batteries in future iterations, the network switch uses a transformer for decoupling.



\subsection{Software Architecture}
Each BeagleBone Black runs Robotic Operating System (ROS) \cite{ros} interface code and actuator control loops in two separate threads. 
The ROS interface code publishes the position, velocity, and effort of the actuators.
Additionally, it subscribes to joint commands sent from a remote computer.
The positions on the U-Joints are regulated using Proportional-Integral-Derivative (PID) controllers. These controllers run on the second thread and update the pulse-width-modulation (PWM) sent to the motor drivers using the magnetic encoder for absolute position feedback. 
The ROS interface thread and controller threads run at 50Hz and 125Hz, respectively.
The remote desktop uses ROS's robot visualizer, RViz, and communicates directly to the battery powered camera, GoPro Session 4, through WiFi.
The software architecture also allows for complete communication over WiFi in future iterations by simply removing the network switch and adding a WiFi dongle to the BeagleBone Blacks.

\section{Experiment and Results}
\begin{figure}
    \center
    \includegraphics[width=0.95\columnwidth]{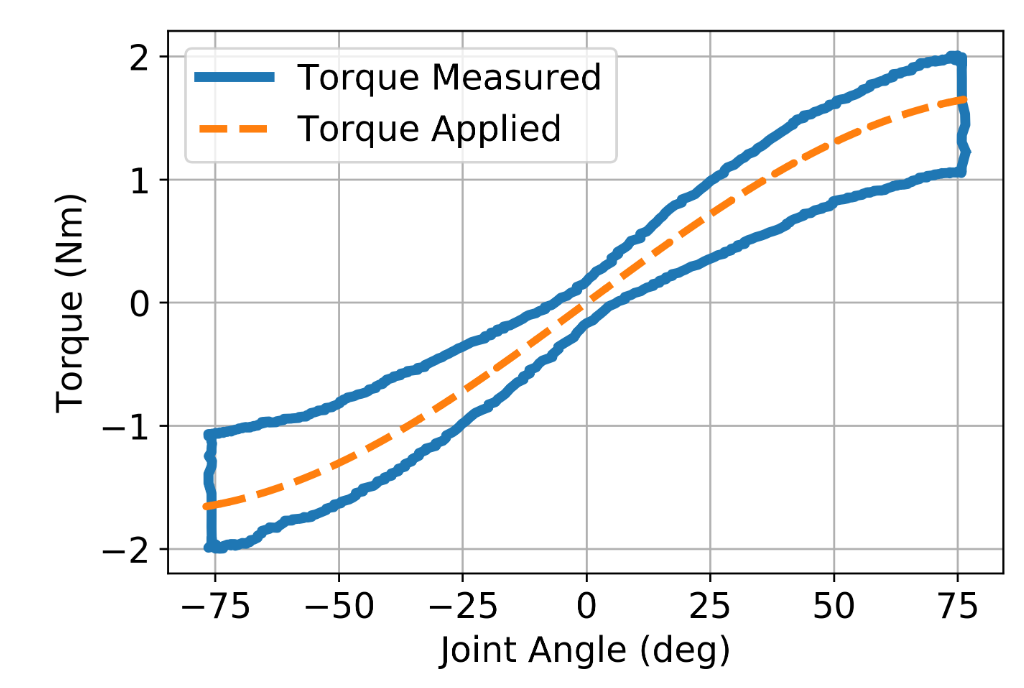}
    \caption{Torque transparency of U-Joint. A cantilevered load is placed on the U-Joint as an \textit{inverted-pendulum-like} configuration. The measured torque is from the onboard sensing of motor currents.}
    \vspace{-5mm}
    \label{fig:transparency}
\end{figure}
\subsection{Configurations for Locomotion}
Three separate configurations (straight, square, and M-shaped) were performed. During this test, the U-Joints are regulated to predefined setpoint positions, then the screw motors are regulated at predefined velocities to induce locomotion.
The configurations are shown in Fig. \ref{fig:gaits}.

\subsection{Torque Transparency}
Joint torque sensing is necessary for active force control. Here, passive sensing of joint force is performed by measuring the motor current. This transmission's transparency is primarily limited by friction, damping, and stiction within the gearbox, as well as motor cogging torque. These effects collectively limit the ability to perform torque control based on motor currents. 

A test, as described in Fig. \ref{fig:transparency}, was performed to evaluate the U-Joint's torque transparency. The test exhibited a maximum hysteresis of 0.96Nm, with a corresponding maximum torque measurement error of 0.62Nm during the change of direction. Therefore, the U-Joint is sufficiently transparent to allow contact detection and force modulation. Additional calibration can further improve the measurement accuracy.

\begin{figure}[t!]
    \vspace{2mm}
    \center
    \includegraphics[width=0.85\columnwidth]{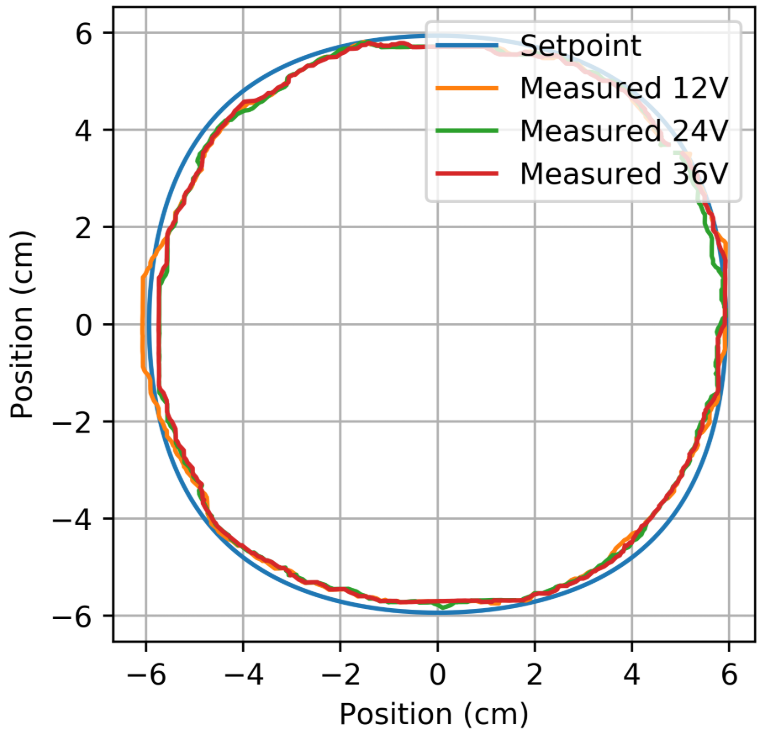}
    \caption{A module is placed in an inverted pendulum configuration. The U-Joint regulates the body position to a planar circular trajectory. The experiment is repeated while varying the voltage inputs into the electrical system. This demonstrates the system's robustness to changing input voltages.}
    \label{fig:regulating_different_voltages}
\end{figure}

\subsection{Additional Experiments}

A single module is placed in an inverted pendulum configuration by clamping the bottom of a U-Joint on a table. The U-Joint is then controlled to move the body in a circle where the off-angle is set to $45^\circ$. The test is repeated with 12V, 24V, and 36V as the input to the cape. Under all three different voltage conditions, the system remains stable and regulates the body's position as seen in Fig. \ref{fig:regulating_different_voltages}. 

\begin{figure}[t!]
\vspace{1mm}
    \center
    \includegraphics[width=0.575\columnwidth]{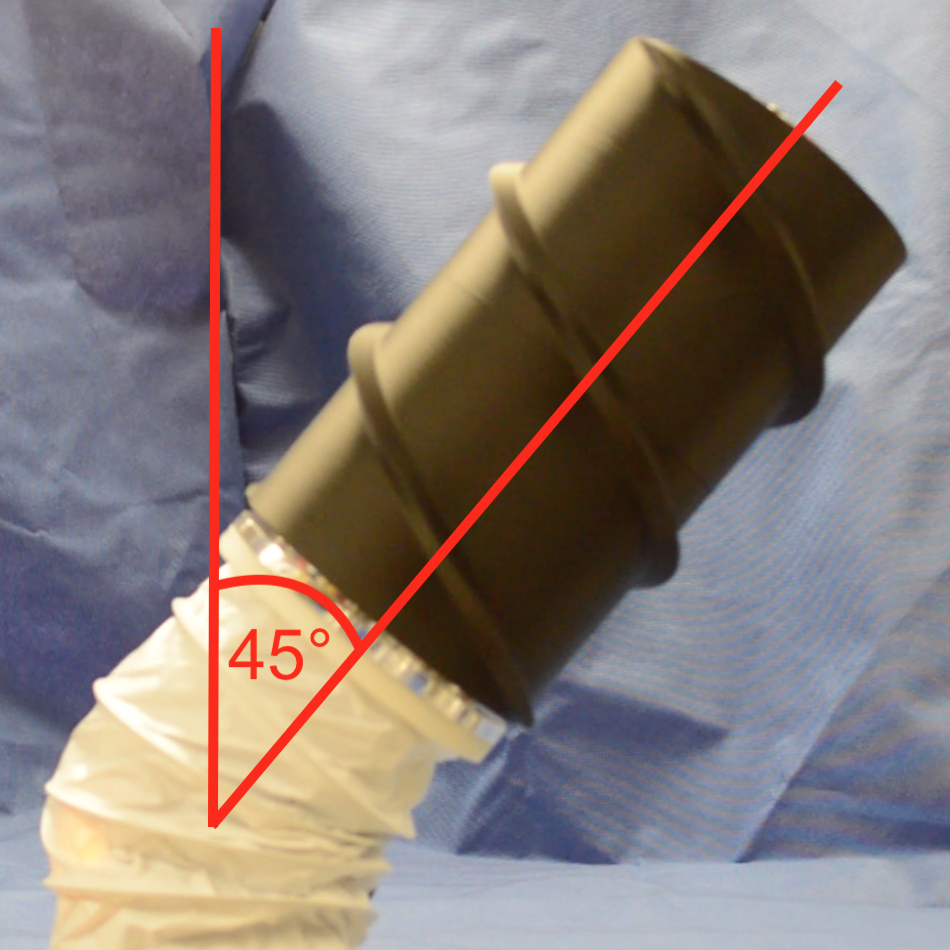}
    \includegraphics[width=0.4\columnwidth]{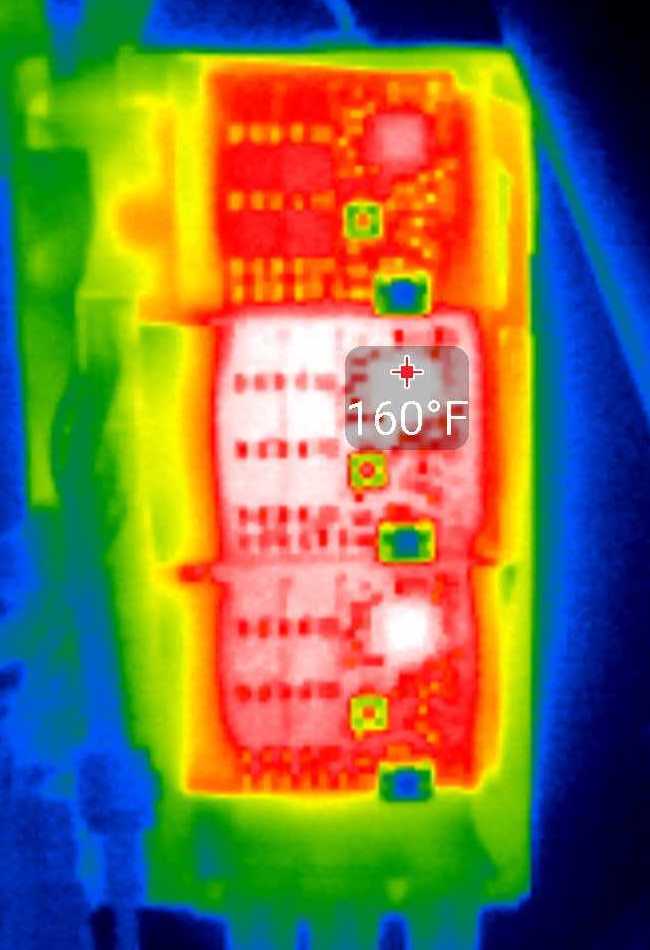}
    \caption{From left to right the figures show a module in an inverted pendulum configuration and a thermal image of the custom cape under high load. The highest temperature recorded is 71$^{\circ}$C.}
    \label{fig:additional_exp}
\vspace{-5mm}
\end{figure}

Separately, a BeagleBone Black with the custom cape is placed outside the module and drives two motors under heavy load. A thermal image is captured and results in a max temperature of 160$^{\circ}$F. Images from both experiments are shown in Fig. \ref{fig:additional_exp}.

\section{Discussion and Conclusion}

Our adaptive system with redundant sensing allows this system to serve as a platform for future explorative development since the control architecture can be changed without hardware reconfiguration. 
This will enable us to explore new methods of locomotion for various environments. For instance, an impedance controller can be implemented by switching the motor controllers to torque mode, and due to the fairly low gear ratio of the motors combined with the transparency of the belt drives, the controller can provide a good approximation for joint torques. 
Since the communication is sent over the network via ROS, a remote desktop can easily be used to control the robot.  This allows computation to be offloaded and integrate higher level autonomy where computation is less constrained

Future work will involve terrestrial exploration and unknown terrain. This will likely require hermetic sealing and tether-free operation through the integration of batteries, wireless communication, and more advanced control computed locally on the robot.

\section{Acknowledgements}
This work was funded by the NASA Jet Propulsion Laboratory under the 2018-2019 Spontaneous Concept Award. The authors would like to thank JPL for assistance in 3D printing of parts and Spencer Chang, Pranay Mehta for help with manufacturing and assembly. D. Schreiber and F. Richter are supported via the National Science Foundation Graduate Research Fellowships.

\balance
\bibliographystyle{ieeetr}
\bibliography{references}

\end{document}